\documentclass{article} 
\usepackage{iclr2021_conference,times}
\usepackage[linesnumbered,ruled]{algorithm2e}
\usepackage{booktabs}
\usepackage{multirow}
\usepackage{xcolor,colortbl}

\usepackage{amsmath,amsfonts,bm}

\def\eqref#1{equation~\ref{#1}}

\def\1{\bm{1}}

\DeclareMathAlphabet{\mathsfit}{\encodingdefault}{\sfdefault}{m}{sl}
\SetMathAlphabet{\mathsfit}{bold}{\encodingdefault}{\sfdefault}{bx}{n}

\newcommand{\R}{\mathbb{R}}

\usepackage{hyperref}
\usepackage{url}
\usepackage{mathtools}
\usepackage[labelfont=bf]{caption} 
\usepackage{makecell}
\title{Unsupervised Meta Learning for One Shot \\Title Compression in Voice Commerce}

\author{Snehasish Mukherjee \\
Department of Computer Science\\
Stanford University\\
\texttt{mukherji@stanford.edu} 
}

\iclrfinalcopy 
\begin{document}

\maketitle

\begin{abstract}
Product title compression for voice and mobile commerce is a well studied problem with several supervised models proposed so far. However these models have 2 major limitations; they are not designed to generate compressions dynamically based on cues at inference time, and they do not transfer well to different categories at test time. To address these shortcomings we model title compression as a meta learning problem where we ask \textit{can we learn a title compression model given only 1 example compression?} We adopt an unsupervised approach to meta-training by proposing an automatic task generation algorithm that models the observed label generation process as the outcome of 4 unobserved processes. We create  parameterized approximations to each of these 4 latent processes to get a principled way of generating random compression rules, which are treated as different tasks. For our main meta learner, we use 2 models; $M_1$ and $M_2$. $M_1$ is a task agnostic embedding generator whose output feeds into $M_2$ which is a task specific label generator. We pre-train $M_1$ on a novel unsupervised segment rank prediction task that allows us to treat $M_1$ as a segment generator that also learns to rank segments during the meta-training process. Our experiments on 16000 crowd generated meta-test examples show that our unsupervised meta training regime is able to acquire a learning algorithm for different tasks after seeing only 1 example for each task. Further, we show that our model trained end to end as a black box meta learner, outperforms non parametric approaches. Our best model obtains a F1 score of 0.8412, beating the baseline by a large margin of 25 F1 points.
\end{abstract}

\section{Introduction}

Voice commerce is one of the fastest growing business segments within e-commerce and is estimated to reach a market size of \$40 Billion by 2022. One of the central problems in designing a good voice shopping experience is to generate voice friendly product titles that are short and can be read out fast. Product titles used in e-commerce are often very long and hence not suitable for use in a voice based conversation. For example, consider the product title ``\textit{OGX Ever Straightening + Brazilian Keratin Therapy Smoothing Shampoo with Coconut Oil, Cocoa Butter \& Avocado Oil for Lustrous, Shiny Hair, Paraben-Free, Sulfate-Free Surfactants, 13 fl.oz}''. Reading out this title will take around 12-15 seconds, degrading the end user experience. Hence this title needs to be compressed to generate a shorter version like ``\textit{OGX Shampoo}'' before it can be used in Voice Shopping. Formally, title compression can be defined as the task that compresses a product title $x=(w_1, w_2, ..., w_n)$ containing $n$ tokens to a shorter one $x_{c} = (w_{c_1}, w_{c_2}, ..., w_{c_k})$ containing $k$ tokens where $k < n$. In this paper we only consider extractive title compression which means that the compressed title is strictly made up of tokens from the input. In other words, $w_{c_i} \in x, i = 1, 2, ..., k$.

Title compression is a well studied problem and several supervised models have been proposed so far. However these models have 2 major limitations. First, given an input product, the compression generated is either always the same or, has a tunable fluency. In either case, the output cannot dynamically adapt to cues provided at inference time. To understand why this might be important, notice that there can be possibly many different compressions for a given product. For example, the product name provided in the previous paragraph can be compressed to ``\textit{OGX Keratin Therapy Shampoo}'' or ``\textit{OGX Shampoo with Coconut Oil}'' etc. The conversational context determines which of these, and many other possible compressions, is appropriate. If the user's previous utterance was ``\textit{search for shampoo with coconut oil}'',  then the compressed title can be ``\textit{OGX Shampoo with Coconut Oil}'', but if the user had said ``\textit{find me a small bottle of keratin shampoo}'', then the compression could have been ``\textit{OGX Keratin Therapy Shampoo , 13 fl.oz}'' and so on. 

The second problem with existing solutions is that title compression trained on a given category does not generalize well to other categories. A model trained on grocery will fail to correctly compress the product name for a television since the distributional semantics of the tokens for a television will be vastly different from that of a shampoo and even pre-training on large amounts of data might not be able to bridge this distribution shift. Adding many training examples for each category might solve this issue, but obtaining large amounts of labeled data is expensive and not scalable to large e-commerce catalogs that contain in excess 200 Million items. To address these problems we model title compression as meta learning problem and explore if we can train a model to learn a new title compression task, given only 1 example compression from that task. For instance, given the example compression of ``Garnier Whole Blends Smoothing Shampoo with Coconut Oil \& Cocoa Butter Extracts, 22 fl. oz.'' to ``Garnier Shampoo , 22 fl.oz.'', we want our model to learn that ``Neutrogena Hydro Boost Gel Moisturizer with Hyaluronic Acid, Hydrating, 1.7 fl oz'' should be compressed to ``Neutrogena Moisturizer , 1.7 fl oz''. 

Following existing methods, we model the problem as a 2-way sequence token classification task where, for each input token $w_j$ the model predicts a label $y_j \in \{0, 1\}$. Tokens labeled \textbf{1} are retained wile the others are dropped, resulting in compressing the product title. To train a model, say $M$, as a meta learner, we provide an example compression as the input along with the test product to get its compression. If $x_{ex}$, $y_{ex}$, and $x_{ts}$ are the example product tokens, binary labels for the example product tokens, and the test product tokens respectively, then $y_{out} = M(x_{ex}, y_{ex}, x_{ts})$ where $y_{out}$ is the label predictions for the test product. We consider title compression for different categories, and different compressions of the same category, as different tasks. To get around the problem of obtaining labeled data, we propose a novel method for automatically constructing title compression tasks by modeling the observed label generation process as the result of 4 unobserved processes. We make parameterized approximations to these processes to generate random labeling rules, which allows us to apply these rules across different products in the sane category in a principled manner.

Our experiments on 16,000 crowd generated 1-shot title compression examples across 4 different held out categories show that our unsupervised meta-training regime using automatically created tasks allows our model to learn new, human generated, title compression tasks only after looking at 1 example from that task. Further, we show that pre-training our model on a segment rank prediction task and training it end-to-end as a black box meta learner outperforms non-parametric meta training approaches.

\section{Related Work}

\subsection{Product Title Compression}
Product title summarization in the context of voice and mobile shopping has been  studied in \citep{DBLP:conf/aaai/WangTQLLSL18, DBLP:conf/cikm/SunJSPOW18,DBLP:journals/corr/abs-1811-04498, DBLP:conf/sigir/XiaoM19, DBLP:conf/aaai/GongLZOLD19, DBLP:conf/sigir/SMukhe11}. Among them, \citep{DBLP:conf/aaai/GongLZOLD19} defined title summarization as a sequence classification problem, where a binary decision is made at each word, given a long product title. Term frequency (tf) and inverse document frequency (idf) for each word has been been used as unsupervised features. However, in this approach large amounts of training data (500,000 samples) have been used and the results didn't prove to be significantly better than a simple BiLSTM approach. Another interesting approach has been proposed, where the problem has been defined as multi-task learning objective \citep{DBLP:conf/aaai/WangTQLLSL18} to compress the product title using user search log data. The multi-task learning objective involves two networks, one network to select the most informative words from the product title as a compressed title and the other network to generate the user search query from the product title. These two networks have been constrained to share the same product title encoder and additionally, the attention distributions from these two networks were constrained to agree with each other. However, in this approach, the size of the training data is again large (185,386 samples) and additionally requires user search data. Another recent approach has been proposed in which the product title summarization has been framed as a Binary Named Entity Recognition problem \citep{DBLP:conf/sigir/XiaoM19}. The model architecture involves a simple bi-directional LSTM encoder/decoder
network (2 layer LSTMs) with an attention mechanism. ANOVA and post-hoc tests showed that with this approach, there was no statistical difference between model outputs and human-labeled short titles. The problem of requiring large amounts of training data was addressed in \cite{DBLP:conf/sigir/SMukhe11} which explores the problem in a low resource setting by discriminatively pre-training a small network to achieve resilience in a low data regime and outperforming models that use 55X more parameters. However none of the proposed solutions address the issue of domain adaptation for few shot title compression.

\subsection{Meta-Learning in NLP}
Meta Learning has shown considerable promise in the area of few shot learning for NLP tasks  \cite{DBLP:journals/corr/abs-2007-09604}. While the focus has mostly been on few shot text classification \cite{DBLP:journals/corr/abs-1806-00852,DBLP:conf/acl/GengLLSZ20} and machine translation \cite{DBLP:conf/emnlp/GuWCLC18}, some recent work \cite{DBLP:journals/corr/abs-2005-14165, DBLP:journals/corr/abs-1911-03863, DBLP:conf/emnlp/DouYA19} have shown the general applicability of meta learning to a wide variety of NLP tasks by improving upon very strong baselines on the GLUE benchmark. Most of the approaches \cite{DBLP:journals/corr/abs-1911-03863, DBLP:conf/emnlp/DouYA19, DBLP:conf/emnlp/GuWCLC18, DBLP:journals/corr/abs-1806-00852} propose some variant of MAML with some interesting modifications being the LEOPARD algorithm \cite{DBLP:journals/corr/abs-1911-03863} that enables optimization based meta learning across tasks with different number of classes and ATAML \cite{DBLP:journals/corr/abs-1806-00852} which separates out task independent representation learning from task specific parameter tuning using attention. Recently \cite{DBLP:journals/corr/abs-2005-14165}, popularly known as GPT-3, has improved upon state-of-the-art results across many different NLP benchmarks in a few-shot setting by training a 175 Billion parameter Transformer language model as a black box meta learner. Though not directly related to NLP,  \cite{DBLP:conf/iclr/HsuLF19} shows that automatic meta training tasks generated from clusters in the embedding space can be used to meta-train a model to acquire a learning algorithm for an actual task after seeing only a few examples. Extending this work, \cite{DBLP:conf/aaai/DengZSCC20} show that for NLP tasks, pre-training the model before meta-training improves the model performance.

\section{Model}

We model the title compression problem as a 2-way sequence token classification problem. If $\mathbf{x}^i = (w_1^i, w_2^i, ..., w_n^i)$, be a product with tokens $n$ tokens $w_1^i,...,w_n^i$, then the binary label vector $\mathbf{y}^i = (y_1^i, y_2^i,..., y_n^i), \text{where } y_j^i \in \{0, 1\}$ induces a title compression on $\mathbf{x}^i$ such that we retain $w_j^i$ if $y_j^i = \mathbf{1}$, else we drop it. A 1-shot meta-training example would then be represented as $\mathcal{D}_{tr}^{i} = (\mathbf{x}_{tr}^i, \mathbf{y}_{tr}^i, \mathbf{x}_{ts}^i, \mathbf{y}_{ts}^i)$ where $\mathbf{x}_{tr}^i$ and $\mathbf{y}_{tr}^i$ are the example product tokens and it's binary labels respectively, and $\mathbf{x}_{ts}^i$ and $\mathbf{y}_{ts}^i$ are the test products and it's labels.

\subsection{Modeling the Label Generation Process}

Each different compression rule is a task. Since we are dealing with 2-way classification, theoretically there exists $2^n - 1$ different compressions for a product with n tokens, corresponding to the number of different non-empty subsets of tokens that can be labeled \textbf{1}. However a random subset selected to be retained in this way does not tell us how to apply the same process on another product in a semantically coherent manner. As in \cite{DBLP:conf/iclr/HsuLF19}, clustering the tokens in an embedding space to compute semantic similarity between tokens could have been used to apply a random compression rule on a different product. However, unlike images, text embeddings are greatly influenced by language modeling task that generated them and are not necessarily good discriminator across tokens of the same product. In other words, while similar images are expected to be close by in the embedding space, there is no intrinsic concept of similarity for words. Semantics and similarity are distributional for text.

To get around this problem, we model the observed label generation process $M_{\theta}$ as the result of 4 unobserved latent processes; $f_{\theta_1}$, $g_{\theta_2}$, $h_{\theta_3}$, and $i_{\theta_4}$, such that 
\begin{equation}
 M_\theta = i_{\theta_4} \cdot h_{\theta_3} \cdot g_{\theta_2} \cdot f_{\theta_1}   
\end{equation}
We describe each of these processes below.

\textbf{Sequence segmentation}, $f_{\theta_1}(\mathbf{x}^i)$: This is a task agnostic process that segments the input sequence $\mathbf{x}^i$ into $k$ contiguous sub sequences $s_1^i, s_2^i, ..., s_k^i$ such that  $\mathbf{x}^i = [s_1^i; s_2^i; ...; s_k^i]$ where $[;]$ denotes the concatenation operator. For example the product ``Neutrogena Hydro Boost Gel Moisturizer with Hyaluronic Acid, Hydrating, 1.7fl oz'' can be segmented into $s_1 = [\text{Neutrogena}]$, $s_2 = [\text{Hydro}, \text{Boost}, \text{Gel}]$, $s_3 = [\text{Moisturizer}]$, $s_4 = [\text{with}, \text{Hyaluronic}, \text{Acid}]$, $s_5 = [ \text{Hydrating}]$ and $s_6 = [\text{17}, \text{fl}, \text{oz}]$. The number of segments and the segments themselves might vary. For example the first segment $s_1$ could have been $[\text{Neutrogens}, \text{Hydro}, \text{Boost}, \text{Gel}]$.

\textbf{Segment mapping}, $g_{\theta_2}(\mathbf{s}^i, c)$: Each segment has a logical role in the product description. In the above example, $s_1 = [\text{Neutrogena}]$ specifies the brand, while $s_6 = [\text{17}, \text{fl}, \text{oz}]$ specifies the pack size. We assume that there are a total of $B$ such segment types but we make no further assumptions about the meaning of each segment. The output of the task specific process $g_{\theta_2}(\mathbf{s}^i, c)$ is a sequence of $k$ labels $\ell_1^i, \ell_2^i, ..., \ell_k^i$, where $\ell_j^i \in \{1, ..., B\}$ is the label of segment $s_j^i$. In other words, $g_{\theta_2}(\mathbf{s}^i, c)$ maps each segment received from $f_{\theta_1}$ to one of these $B$ labels. $g_{\theta_2}$ is task specific because the second parameter $c$ can be regarded as the category information. This is important as an electronics product will have different semantics for some of the segments than a grocery product. $g_{\theta_2}$ can be viewed as a NER task, or more appropriately a POS tagging task, but unlike NER and POS tagging, we do not know what entities or parts-of-speech we are looking for. \textit{Note that within a given category $c$, $g_{\theta_2}$ can let let us map the different segments of 2 products thereby giving a direct way of applying a compression rule in a semantically coherent manner across 2 different products from the same category}. 

\textbf{Segment retention}, $h_{\theta_3}(\mathbf{s}^i, \mathbf{\ell}^i, c)$: This task specific process determines which segments to keep and which to drop by generating $k$ binary outputs $r_1^i, r_2^i, ..., r_k^i$ such that segment $s_j^i$ is retained if its retention label $r_j^i = $ \textbf{1}, otherwise it's dropped. 

\textbf{Intra-segment retention}, $i_{\theta_4}(\mathbf{s}^i, \mathbf{\ell}^i, \mathbf{r}^i, c)$: This task specific process determines which tokens to keep within each segment. For each token $w_p$ of each segment $s_j$, it generates a binary label $m_{pj} \in \{0, 1\}$. It then concatenates $m_{pj}$ masked by $r_j$ for all the segments gives rise to the observed labels $\mathbf{y}^i$ for the product $\mathbf{x}^i$. Succinctly, 
\begin{equation}
    \mathbf{y}^i = \text{concat}(m_{pj} * r_{j}), \forall p = 1, 2, ... n_{j}, \forall j = 1, 2, ,,, k
\end{equation}
where $n_j$ is the number of tokens in the segment $s_j$.

\subsection{Unsupervised Meta Training Example Generation}

To automatically generate meta-training examples we create parameterized approximations to each of the 4 latent process $f_{\theta_1}$, $g_{\theta_2}$, $h_{\theta_3}$, and $i_{\theta_4}$. 

\subsubsection{Segment Generation}
For the sequence segmentation, we first train an unsupervised Modified-Kneser-Ney (abb. MKN) smoothed trigram language model \citep{DBLP:conf/acl/ChenG96} on 12 Million product search queries. The choice of the language model is not central to this algorithm; we choose MKN since it is extremely fast to train even on CPU. Then we greedily grow segments starting from the beginning, based on maximizing the length normalized negative log likelihoods of the segments. The parameters for this algorithm are $\theta_1 = (\alpha, t)$, where $\alpha$ is a length normalization parameter and $t$ is the threshold that determines when a new segment should start. The process is formally presented in Algorithm \ref{algo:seg}. Note that, though we use negative log likelihood for numerical stability, Algorithm \ref{algo:seg} uses the raw probabilities, denoted by $P_{MKN}$ for notational simplicity.

\begin{algorithm}
    \SetKwInOut{Input}{Input}
    \SetKwInOut{Output}{Output}

    \Input{Product tokens $\mathbf{x}^i = (w_1^i, w_2^i, ..., w_n^i)$}
    \Output{Segment labels: $\mathbf{s}^i[w_j]$, $j=1, 2, ..., n$}
    $\mathbf{s}[w_1] \gets 1$\\
    \For{$i\gets2$ \KwTo $n$}{
        $\text{left2} = 2^{\alpha} \cdot P_{MKN}(w_{i-1}, w_i)$\\
        $\text{left3} = 3^{\alpha} \cdot P_{MKN}(w_{i-2}, w_{i-1}, w_i)$ if $i >= 3$ else None\\
        $\text{det} = P_{MKN}(w_i) - \text{max}(\text{left2}, \text{left3})$\\
        \eIf{$\text{det} <= \theta_1$}{
            $\mathbf{s}[i] \gets \mathbf{s}[i - 1]$
        }{
            $\mathbf{s}[i] \gets 1 + \mathbf{s}[i - 1]$
        }
        
    }
    \caption{Sequence Segmentation: $f_{\theta_1}$}
    \label{algo:seg}
\end{algorithm}

\subsubsection{Segment Mapping}

Mapping of the segments to one of the $B$ buckets needs to depend primarily on the segment semantics. The segment embeddings can be treated as an approximation of the segment semantics, though as noted earlier, embeddings for text are not always intuitive and might not be a good discriminator across different segments. We train a replaced token detection model \citep{DBLP:conf/iclr/ClarkLLM20}, $E_{rt}$, on 250,000 product titles and pass product tokens through it to generate it's embeddings. The centroid of the token emebddings for a given segment, projected on 2 dimensions through a parameter vector $\theta_a \in \R^{d \times 2}$ where $d$ is the out dimension size from $E_{rt}$, is treated as the segment embedding. We augment the segment embedding by adding 2 more features; the segment likelihood as determined by $P_{MKN}$ and the segment position. We project this 4D vector on to 1D using the parameter vector $\theta_b \in \R^{4 \times 1}$. The ordering of the segments then give us our segment maps $\mathbf{\ell}^i$. Random choices of the parameters $\theta_2 = (\theta_a, \theta_b)$ gives us different orderings of the segments, and hence different segment labels. Algorithm \ref{algo:map} is a formal description of this algorithm.

\begin{algorithm}
    \SetKwInOut{Input}{Input}
    \SetKwInOut{Output}{Output}

    \Input{Product tokens $\mathbf{x}^i = (w_1^i, w_2^i, ..., w_n^i)$ and segments  $\mathbf{s}^i$}
    \Output{Segment maps: $\mathbf{\ell}^i[w_j]$, $j=1, 2, ..., n$}
    $a\gets[]$\\
    \For{$c\gets1$ \KwTo $k$}{
        $a[c]=\{w_j: \mathbf{s}^i[j] = c\}$
    }
    $v1 \gets []$\\
    $v2 \gets \{\}$\\
    \For{$c\gets1$ \KwTo $k$}{
        $n_j \gets \text{len}(a[c])$\\
        $\text{centroid} = \frac{1}{n_j}\sum_{j=1}^{n_j} E_{rt}(a[c][j])$\\ 
         $\text{centroid2D} = \text{centroid}^{1 \times d} \times \theta_a^{d \times 2}$\\
         $\text{f} = [\text{centroid1D}; n_j^{\alpha} \cdot P_{MKN}(a[c]); c]$\\
         $\text{f1D} = f^{1 \times 4} \times \theta_b^{4 \times 1}$\\
         $\text{insertIdx} = \text{binarySearch}(v, \text{f1D})$
         $v1[\text{insertIdx}] \gets f1D$\\
         $v2[c] \gets \text{insertIdx}$
    }
    
    $\ell \gets []$\\
    \For{$j \gets 0$ \KwTo $n$}{
        $\text{segmentId} \gets \{c: w_j \in a[c]\}$\\
        $\ell^i[w_j] \gets v2[\text{segmentId}]$
    }
    \caption{Segment Mapping: $g_{\theta_2}$}
    \label{algo:map}
\end{algorithm}

\subsubsection{Segment Retention and Propagation}

For segment retention we keep the segment with maximum likelihood, and for the rest we randomly generate a binary label. For the segments retained, we generate the retention labels to randomly pick the most likely prefix, suffix or sub string.

These 4 approximations to the latent label generation process allows us to apply a randomly generated compression rule for a given product to another product in a principled and semantically coherent manner.

\subsection{Modeling the Meta Learner}

For our meta learner, we factor $M_\theta$ in to 2 components; $M1_{\theta_1}$ and $M2_{\theta_2}$ and re-write equation 1 as

\begin{equation}
    M_{\theta} = M_{2_{\theta_2}} \cdot M_{1_{\theta_1}}
\end{equation}

Drawing analogy to our treatment of the label generation process in section 3.1, we can treat  $M_1$ as responsible for sequence segmentation and segment mapping, while $M_2$ as responsible for task specific label generation. However, if $M_1$  generates and maps the segments, the optimization of $M_1$ can become more challenging since $M_2$ will then be receiving only categorical output from $M_1$ and information might be lost. So, instead of letting $M_1$ actually output labeled segments, we take the embeddings generated by $M_1$ and directly feed it into $M_2$ and train the model end-to-end. 

We choose a 2 layer Transformer \citep{DBLP:conf/nips/VaswaniSPUJGKP17} with 8 attention heads and 512 hidden units as $M_1$. The input to the Transformer comes from a hybrid embedding layer that uses a highway network to combine 300 dimensional pre trained word2vec embeddings with a 212 dimensional word embeddings derived from character level convolutions. For this embedding layer we use the same implementation as in \cite{DBLP:conf/sigir/SMukhe11}. 

For $M_2$, we have 2 choices; non-parametric and parametric. When using the non-parametric approach, we use a 2-way Prototypical Network \citep{DBLP:conf/nips/SnellSZ17} to compute the output labels based on the proximity of test tokens to the centroids of the example tokens belonging to class \textbf{0} and class \textbf{1}. While training the network end-to-end as a black box meta learner, we use a single headed dot product attention for $M_2$ as in \cite{DBLP:conf/nips/VaswaniSPUJGKP17}, with the modification that the query and value vectors are obtained from 2 different projections of the test product, while the key vector is obtained by convolving the example product with the example labels. The idea here is for $M_2$ to learn to attend to the relevant tokens of the example sequence while computing the labels for the test sequence. 

We summarize our model formally in the following set if equations
\begin{eqnarray}
    \mathbf{e}_{tr}&=&\text{HybridEmbedding}(\mathbf{x}_{word_{tr}}, \mathbf{x}_{char_{tr}}), \mathbf{e}_{tr} \in \R^{bs \times sl \times dim}\\
    \mathbf{e}_{ts}&=&\text{HybridEmbedding}(\mathbf{x}_{word_{ts}}, \mathbf{x}_{char_{ts}}), \mathbf{e}_{ts} \in \R^{bs \times sl \times dim}\\
    \mathbf{r}_{tr}&=&\text{Transformer}(\mathbf{e}_{tr}), \mathbf{r}_{tr} \in \R^{bs \times sl \times dim}\\
    \mathbf{r}_{ts}&=&\text{Transformer}(\mathbf{e}_{ts}), \mathbf{r}_{ts} \in \R^{bs \times sl \times dim}\\
    \mathbf{q}_{ts}&=&\text{Conv1D}(\mathbf{r}_{ts}), \mathbf{q}_{ts} \in \R^{bs \times sl \times dim}\\
    \mathbf{v}_{ts}&=&\text{Conv1D}(\mathbf{r}_{ts}), \mathbf{v}_{ts} \in \R^{bs \times sl \times dim}\\
    \mathbf{k}_{tr}&=&\text{Conv1D}(\mathbf{r}_{tr} \star \mathbf{y}_{tr}), \mathbf{k}_{tr} \in \R^{bs \times sl \times dim}\\
    \mathbf{y}_{pr}&=&M_2(\mathbf{r}_{tr}, \mathbf{r}_{ts}, \mathbf{y}_{tr}), \mathbf{y}_{ts} \in \R^{bs \times sl \times 2}\\
    M_2&=& 
\begin{dcases}
    \text{DotProductAttention}(\mathbf{q}_{ts}, \mathbf{v}_{ts}, \mathbf{k}_{tr}),& \text{if } \text{parametric}\\
    \text{ProtoNet}(\mathbf{r}_{tr}, \mathbf{r}_{ts}, \mathbf{y}_{tr}),              & \text{if } \text{non-parametric}
\end{dcases}
\end{eqnarray}
where $bs$ is the batch size, $sl$ is the maximum sequence length, $dim$ is the hidden unit size and $\mathbf{y}_{pr}$ is the predicted labels. The network is trained to minimize the weighted binary cross entropy loss
\begin{equation}
  L(\theta)=-\frac{1}{bs \cdot sl} \sum_{i = 1}^{bs \cdot sl} \alpha \cdot \mathbf{y}_{ts}\log \left(\mathbf{y}_{pr}\right)+\beta \cdot (1 - \mathbf{y}_{ts})\log (1 - \mathbf{y}_{pr})
\end{equation}
where $\alpha = 0.25$ and $\beta = 1 - \alpha$, since roughly 25\% of the tokens per product are labeled \textbf{1}.

\subsection{Pre-Training}

Most meta-learning approaches, particularly non-parametric ones, operate in the embedding space. As such, the success of a meta-learner is contingent upon the quality of embeddings learned by the base model, $M_1$ in our case. Particularly, we want $M_1$ to learn embeddings that can help discriminate between the different segments of a product, instead of being able to discriminate between products. In other words, instead of tokens belonging to similar products being clustered together in the embedding space, we want the segments that play the same role in product titles to be closer together in the embedding space resulting in clusters that map to segment types. 

In order to achieve this we propose a novel segment rank prediction based pre training task for $M_1$. Given an input product $x = (w_1, w_2, ..., w_n)$, $M_1$ has to predict a label $y_i \in \{1, 2, ..., B\}$ for each token $w_i$. Tokens that have the same label form a segment, i.e.  $s_i = \{w_j: y_j = i\}, i \in \{1, 2, ..., k\}, k \leq B$, and the label itself is indicative of the rank of the segment, so that $rank(s_i) = i$. To enable $M_1$ make the categorical prediction, we attach a time distributed softmax classifier with $B$ hidden units to $M_1$.

To generate unsupervised training data for this task, we segment a long product titles following Algorithm \ref{algo:seg}, and we rank each segment based on the length normalized probability of observing that segment following step 2 of Algorithm \ref{algo:seg}, except now we compute the probability of the entire segment instead of bigrams. The intuition here is to lend a sufficiently strong prior to $M_1$ that allows it to recognize which tokens are grouped together and which token groups are important. Ranking segments like this can be regarded as a method of matching different segments across 2 different products. We apply this algorithm on 250,000 products to automatically generate training data and train $M_1$ for 30 epochs on categorical cross entropy loss.

\section{Experiments}

In this section we first describe the dataset used and then describe our experiments and the research questions they are designed to answer.

\subsection{Dataset}
We use 5 private datasets for training different models. Following is a brief description of each dataset and how it has been used. 

$\mathcal{D}_{search}$ is a dataset consisting of the top 12 Million product search queries on Walmart.com and their frequencies over a 1 year period. We use the distribution induced by the length normalized frequencies of the queries to sample 1 Million distinct queries from this dataset to train a modified-Kneser-Ney smoothed trigram language model, $P_{MKN}$, introduced in Section 3. $\mathcal{D}_{product}$ is a dataset of 250,000 top selling Walmart products over the last 6 months. This dataset is used to pre-train $M_1$ on a replaced token detection task \citep{DBLP:conf/sigir/SMukhe11}. $\mathcal{D}_{com-human}$ is a dataset of 40,445 human generated title compressions spanning 8 different product categories in the Walmart catalog. Each row in this dataset is structured as $(x, y)$ where $x$ is the product tokens and $y$ is the binary token labels. In this dataset, each product appears only once, i.e. there is only 1 compression example per product. We use this dataset to fine tune $M_1$.

$\mathcal{D}_{meta-auto}$ and $\mathcal{D}_{meta-human}$ are our main datasets. $\mathcal{D}_{meta-auto}$ is the dataset consisting of 40,000 meta training example generated automatically using the Algorithm in section 3.2. 1000 random product pairs were sampled from each of 4 categories, and 4 compressions were generated for each pair to get 40,000 rows of meta training data. Each row  of this dataset is structured as $(x_{ex}, y_{ex}, x_{ts}, y_{ts})$ where the first tuple is a product compression used as meta-training example, while the second tuple is a similar compression used as the meta-test. $\mathcal{D}_{meta-human}$ is a dataset consisting of 16,000 human generated 1-shot title compression examples with each row having the exact same structure as $\mathcal{D}_{meta-auto}$. This dataset covers 1000 products from each of 4 held out categories, i.e. categories that are different from the ones used to generate $\mathcal{D}_{meta-auto}$. This is the dataset used for all our evaluations. Note: $\mathcal{D}_{meta-human}$ evaluates models both on capability to generate different compressions for the same product and capability to adapt to unseen classes at test time.

\subsection{Methodology}

All datasets were uniformly normalized and converted to lower case, ampersands (\&) were converted to ``and'', and hyphens (i.e. - ) were dropped. No other special characters were removed. We followed a simple tokenization scheme of splitting strings by space character. All training, including pre-training, fine tuning and meta training, were carried out for 30 epochs. All trainings used Adam optimizer with the learning rate fixed at 0.0001. We did not tune any other hyperparameters. For evaluation we report both the F1 scores and the exact match (EM) scores wrt $(x_{ts}, y_{ts})$ of $\mathcal{D}_{meta-human}$.

\subsection{Experiment Design}

In this section we describe our experiments and the research questions they are designed to answer.  

\textbf{Benefit of Meta-Learning}: Is meta-learning at all necessary for title compression? To explore this question we take 2 baseline models. The first model is CB3SA+PT from \cite{DBLP:conf/sigir/SMukhe11}, which is finetuned on $\mathcal{D}_{com-human}$. To maintain notational uniformity we call this model CB3SA+rePT. The second model is $M_1$ with a point wise binary softmax classifier attached as the outer layer. We call this model M1+rePT. It is also pre-trained and finetuned exactly like CB3SA+rePT. To attain parity with our meta learner w.r.t. seeing the example data during inference, we finetune both these models for 1 more epoch on the examples $(x_{ex}, y_{ex})$ in $\mathcal{D}_{com-human}$. We compare these 2 baselines against our proposed model introduced in Section 3.3, with both with parametric and non-parametric implementations of $M_2$, both of them pre-trained on replace token detection task. We name these models BB+rePT and PN+rePT respectively. 

\textbf{Benefit of segment rank prediction based pre-training}: We compare 3 different pre-training strategies; no pre training (noPT), replaced token detection pre training (rePT) and segment rank prediction based pre training (raPT). We pre-train all 3 models; M1, BB and PN, with all 3 pre-training strategies and record the performance in section 2 of Table 1.

\textbf{Task construction ablations}: We experiment with 2 methods of constructing meta training tasks; the method introduced in Section 3.2 which we code name SL (as in segment labeling), and clustering in embedding space, which we code name EC (as in embedding cluster). For EC, our method is similar to that of \cite{DBLP:conf/iclr/HsuLF19}. We use k-means clustering on the embeddings of tokens generated by M1+rePT to obtain $B$ clusters. Each task is constructed by randomly selecting a product, computing it's token cluster ids, and then selecting another product from the same category that has a super set of these clusters. Then we randomly select a subset of these common clusters to retain. We evaluate performance of both BB and PN, pretrained on meta-learning tasks generated by SL and EC.

\section{Results and Discussion}

The results of our experiments are tabulated in Table 1. Note that all rows are not distinct since some of them have been repeated for easy comparison. Naming convention of the models are as defined in Section 4.3.

\begin{table}[h]
  \setlength\extrarowheight{4pt}
  \centering
  \caption{Performance on test set  $\mathcal{D}_{meta-human}$ }
  \label{tab:ablation}
  \begin{tabular}{p{0.15\textwidth}p{0.35\textwidth}p{0.15\textwidth}p{0.15\textwidth}}
    \toprule
    \textbf{Experiment}&\textbf{Model}&\textbf{F1}&\textbf{EM}\\
    \midrule
    \multirow{4}{*}{Meta-training}&
        \multicolumn{1}{l}{CB3SA+rePT}& 
        \multicolumn{1}{l}{0.5513} & 
        \multicolumn{1}{l}{5.31}\\
        \cline{2-4}
        &\multicolumn{1}{l}{M1+rePT}& 
        \multicolumn{1}{l}{0.5897} & 
        \multicolumn{1}{l}{7.06}\\
        \cline{2-4}
        &\multicolumn{1}{l}{PN+rePT}& 
        \multicolumn{1}{l}{0.5985} & 
        \multicolumn{1}{l}{12.19}\\
        \cline{2-4}
        &\multicolumn{1}{l}{BB+rePT}& 
        \multicolumn{1}{l}{\cellcolor{green!25}0.7841} & 
        \multicolumn{1}{l}{\cellcolor{green!25}16.0}\\
        
    \midrule
    
    \multirow{9}{*}{Pre-training}&
        \multicolumn{1}{l}{M1+noPT}& 
        \multicolumn{1}{l}{0.4914} & 
        \multicolumn{1}{l}{5.37}\\
        \cline{2-4}
        &\multicolumn{1}{l}{M1+rePT}& 
        \multicolumn{1}{l}{0.5897} & 
        \multicolumn{1}{l}{7.06}\\
        \cline{2-4}
        &\multicolumn{1}{l}{M1+raPT}& 
        \multicolumn{1}{l}{0.6111} & 
        \multicolumn{1}{l}{7.92}\\
        \cline{2-4}
        &\multicolumn{1}{l}{PN+noPT}& 
        \multicolumn{1}{l}{0.5156} & 
        \multicolumn{1}{l}{9.81}\\
        \cline{2-4}
        &\multicolumn{1}{l}{PN+rePT}& 
        \multicolumn{1}{l}{0.5985} & 
        \multicolumn{1}{l}{12.19}\\
        \cline{2-4}
        &\multicolumn{1}{l}{PN+raPT (ours)}& 
        \multicolumn{1}{l}{0.7030} & 
        \multicolumn{1}{l}{30.18}\\
        \cline{2-4}
        &\multicolumn{1}{l}{BB+noPT}& 
        \multicolumn{1}{l}{0.6288} & 
        \multicolumn{1}{l}{0.00}\\
        \cline{2-4}
        &\multicolumn{1}{l}{BB+rePT}& 
        \multicolumn{1}{l}{0.7841} & 
        \multicolumn{1}{l}{16.0}\\
        \cline{2-4}
        &\multicolumn{1}{l}{BB+raPT (ours)}& 
        \multicolumn{1}{l}{\cellcolor{green!25}0.8412} & 
        \multicolumn{1}{l}{\cellcolor{green!25}35.75}\\

    \midrule
    
    \multirow{4}{*}{Task Creation}&
        \multicolumn{1}{l}{PN+EC}& 
        \multicolumn{1}{l}{0.4581} & 
        \multicolumn{1}{l}{3.55}\\
        \cline{2-4}
        &\multicolumn{1}{l}{PN+SL (ours)}& 
        \multicolumn{1}{l}{0.7030} & 
        \multicolumn{1}{l}{30.18}\\
        \cline{2-4}
        &\multicolumn{1}{l}{BB+EC}& 
        \multicolumn{1}{l}{0.4830} & 
        \multicolumn{1}{l}{5.03}\\
        \cline{2-4}
        &\multicolumn{1}{l}{BB+SL (ours)}& 
        \multicolumn{1}{l}{\cellcolor{green!25}0.8412} & 
        \multicolumn{1}{l}{\cellcolor{green!25}35.75}\\
  \bottomrule
\end{tabular}
\end{table}

\textbf{Benefit of Meta-Learning:} The first section of Table 1 compares 2 baselines against our meta-learners. Clearly the black box meta learner (BB+rePT) outperforms the baselines, but interestingly our non-parametric meta learner (PN+rePT) barely beats the best performing baseline by 1 F1 point. This can be attributed to the fact that non-parametric approaches are reliant of the quality of embeddings learnt by the base layer, which in turn, depends on the pre-training algorithm. For this comparison we used the replaced token detection based pre-training \citep{DBLP:conf/iclr/ClarkLLM20, DBLP:conf/sigir/SMukhe11} which, as discussed later, is not a good discriminator between token types.

\textbf{Benefit of segment rank prediction based pre-training:} The second section of the table shows that all models benefit from pre-training (noPT vs *PT), and our proposed segment rank prediction based pre-training (raPT) out performs replaced token detection based pre-training (rePT). These observations are in line with \cite{DBLP:conf/aaai/DengZSCC20} which shows that pre-training before meta-training almost always improves performance. Replaced token detection based pre-training underperforms our proposed segment rank based pre-training, because in rePT, replacement tokens are drawn from a skip gram distribution which causes the model to learn token distributions over a window, instead of adjacent token distribution as in raPT. As a result, for rePT, tokens belonging to the same product, or product sub category would form a cluster in the embedding space. In contrast, with raPT we explicitly train the network to recognize and discriminate between segments, so tokens of the same type of segments (as determined by their ranks) would cluster together. This makes a more difference for PN than BB, since PN directly operates on clusters in the embedding space.

\textbf{Task Construction Ablations}: Task creation based on clusters in the embedding space (EC) performs even worse than no pre-training at all (*+PT). The reasons are same as that presented in the section above. Since the clusters in the embedding space do not discriminate very well between tokens of the same product or subcategory, the examples do not have lot of variance. Also, unlike our proposed task generation algorithm, there is no guarantee of spatial coherence in the cluster members. As a result, generated tasks often include skip tokens that do not map well to the actual token semantics. 

\section{Conclusion and Future Work}

We model title compression as a meta-learning problem and propose an automatic way of generating meta-training examples. We show that our meta learner, meta-trained on our synthetic examples is able to learn a model for actual human generated title compression tasks, by seeing only 1 example. We also propose a novel segment rank prediction based pre-training task which is shown to further improve performance on the test set. Our best model, a black box meta learner pre-trained on the segment rank prediction task, outperforms a strong baseline by nearly 25 F1 points. A promising research direction from here would be to explore how we can leverage MAML to optimize $M_1$ at test time and if using MAML removes the necessity of pre-training the base embedding generator. 

\clearpage



\bibliography{iclr2021_conference}
\bibliographystyle{iclr2021_conference}

\end{document}